\documentclass{article}

\usepackage{microtype}
\usepackage{graphicx}
\usepackage{subcaption}
\usepackage{booktabs} % for professional tables

\usepackage{hyperref}

\newcommand{\SCP}{\textsc{SA-MSCP}} 

 \usepackage[preprint]{icml2026}

\usepackage{amsmath}
\usepackage{amssymb}
\usepackage{mathtools}
\usepackage{amsthm}

\usepackage[capitalize,noabbrev]{cleveref}

\theoremstyle{plain}

\theoremstyle{definition}

\theoremstyle{remark}

\usepackage[textsize=tiny]{todonotes}

\icmltitlerunning{Simulation-Augmented MSCP for Aggregated Forecasts}

\begin{document}

\twocolumn[
  \icmltitle{Simulation-Augmented Multi-Step Split Conformal Prediction for Aggregated Forecasts}

  % It is OKAY to include author information, even for blind submissions: the
  % style file will automatically remove it for you unless you've provided
  % the [accepted] option to the icml2026 package.

  % List of affiliations: The first argument should be a (short) identifier you
  % will use later to specify author affiliations Academic affiliations
  % should list Department, University, City, Region, Country Industry
  % affiliations should list Company, City, Region, Country

  % You can specify symbols, otherwise they are numbered in order. Ideally, you
  % should not use this facility. Affiliations will be numbered in order of
  % appearance and this is the preferred way.
  \icmlsetsymbol{equal}{*}

  \begin{icmlauthorlist}
    %\icmlauthor{Firstname1 Lastname1}{equal,yyy}
    %\icmlauthor{Firstname2 Lastname2}{equal,yyy,comp}
    \icmlauthor{Andro Sabashvili}{comp}
    %\icmlauthor{Firstname4 Lastname4}{sch}
    %\icmlauthor{Firstname5 Lastname5}{yyy}
    %\icmlauthor{Firstname6 Lastname6}{sch,yyy,comp}
    %\icmlauthor{Firstname7 Lastname7}{comp}
    %\icmlauthor{}{sch}
    %\icmlauthor{Firstname8 Lastname8}{sch}
    %\icmlauthor{Firstname8 Lastname8}{yyy,comp}
    %\icmlauthor{}{sch}
    %\icmlauthor{}{sch}
  \end{icmlauthorlist}

  %\icmlaffiliation{yyy}{Department of XXX, University of YYY, Location, Country}
  \icmlaffiliation{comp}{Andro Sabashvili}
  %\icmlaffiliation{sch}{School of ZZZ, Institute of WWW, Location, Country}

  %\icmlcorrespondingauthor{Firstname1 Lastname1}{first1.last1@xxx.edu}
  \icmlcorrespondingauthor{Andro Sabashvili}{andro.sabashvili@gmail.com}

  % You may provide any keywords that you find helpful for describing your
  % paper; these are used to populate the "keywords" metadata in the PDF but
  % will not be shown in the document
  \icmlkeywords{Conformal prediction, Time-series forecasting, Aggregated forecasts, Uncertainty quantification}

  \vskip 0.3in
]

% this must go after the closing bracket ] following \twocolumn[ ...

% This command actually creates the footnote in the first column listing the
% affiliations and the copyright notice. The command takes one argument, which
% is text to display at the start of the footnote. The \icmlEqualContribution
% command is standard text for equal contribution. Remove it (just {}) if you
% do not need this facility.

% Use ONE of the following lines. DO NOT remove the command.
% If you have no special notice, KEEP empty braces:
\printAffiliationsAndNotice{}  % no special notice (required even if empty)
% Or, if applicable, use the standard equal contribution text:
% \printAffiliationsAndNotice{\icmlEqualContribution}

\begin{abstract}
  We study uncertainty quantification for aggregated forecasting tasks such as annual totals and year-over-year growth rates. We propose \SCP{}, a simulation-augmented multi-step split conformal method that generates future paths from cross-validated residuals using a block bootstrap and constructs prediction intervals from empirical quantiles. Experiments show that \SCP{} improves empirical coverage over a simulated-path baseline for aggregated and growth-rate targets. Our results demonstrate that simulation-enhanced conformal calibration is an effective and general framework for uncertainty quantification in aggregated time-series forecasting. 
\end{abstract}

\section{Introduction}  
Prediction intervals for time-series forecasts are central to applications such as retail and finance. Conformal prediction (CP) provides distribution-free coverage under exchangeability \citep{Vovk2005,Lei2018}. Extensions address temporal dependence \citep{XuXie2021,XuXie2023,WangHyndman2024}, online adaptation \citep{Gibbs2021,Angelopoulos2023}, weighted quantiles \citep{tibshirani2019conformal,Barber2023}, time-series datasets \citep{Stankeviciute2021,SunYu2022}, and distribution shifts \citep{zou2024coverage}.  

Many applications require uncertainty for aggregated targets such as annual sums and year-over-year growth rates. This is challenging because aggregation induces dependence and nonlinear transformations, so combining pointwise intervals is invalid, and CP guarantees do not directly extend to multi-step dependent settings.  

We propose \SCP{}, a simulation-enhanced multi-step split conformal method \citep{WangHyndman2024} for aggregated forecasting. The method uses expanding-window cross-validation to collect residuals across forecast origins, then simulates future paths with a block bootstrap over residual sequences to better preserve dependence before constructing prediction intervals from empirical quantiles of aggregated trajectories.  

Related work includes copula-based calibration for joint coverage \citep{SunYu2022}, conformalized interval arithmetic for sums under group exchangeability \citep{LuoZhou2025}, and hierarchical reconciliation approaches \citep{Principato2025}. In contrast, \SCP{} focuses on temporal aggregation within a univariate series.  

Our contributions are: (i) a practical MSCP-based procedure for aggregated targets that improves empirical coverage without finite-sample guarantees, and (ii) an empirical evaluation on M4 and a proprietary dataset of 2{,}000 series showing consistent coverage gains over a simulated-path baseline with significance assessed via Wilcoxon tests.  

\section{Methodology}  
Split conformal prediction (SCP) constructs prediction intervals from calibration residuals and achieves marginal coverage under exchangeability \citep{Lei2018,ShaferVovk2008}. In time series, temporal dependence violates this assumption, motivating multi-step conformal approaches that approximate residual distributions via rolling or expanding windows.

\SCP{} follows this framework. Residuals $\hat{\varepsilon}_{k,h} = y_{k+h} - \hat{y}_{k+h}$ are collected via expanding-window cross-validation across forecast origins $k$ and horizons $h$. In our implementation, the cross-validation uses an expanding window with an initial calibration window of $10$ observations; at each step the window grows by one observation and the next $h$ points are used for validation. We center each horizon-wise residual column, extract all valid consecutive residual blocks of length $b$, and simulate $S$ future paths by resampling these blocks with replacement and stitching them together along the horizon. This block bootstrap preserves local residual dependence better than horizon-wise independent resampling.  

In this paper, we apply the method to monthly time series with annual aggregates, but the same approach can be used with other temporal resolutions and aggregation units. Each simulated path yields monthly forecasts, which are aggregated to annual totals $\widehat{Y}_{s} = \sum_{m=1}^{12} \widehat{y}_{s,m}$ and transformed into growth rates using consecutive annual values. Prediction intervals are then obtained from empirical quantiles of the simulated samples at the desired coverage levels for both pointwise and aggregated targets.  

We use block size $b=12$ for M4, matching its 36-month forecast horizon and preserving annual dependence patterns, and $b=3$ for the proprietary data. The horizon is only 12 months for the proprietary data, so using $b=12$ would leave too few distinct bootstrap combinations; $b=3$ instead preserves within-quarter dependence while retaining sufficient resampling diversity. This simulation-based calibration approximates the distribution of aggregated quantities under temporal dependence, enabling uncertainty quantification beyond pointwise horizons.
  
\begin{algorithm}[tb]  
  \caption{\SCP{}: Simulation-Enhanced SCP for Aggregated Forecasts}  
  \label{alg:scp}  
  \begin{algorithmic}[1]  
    \REQUIRE Training series $y_{1:T}$, test start $T{+}1$, expanding-window CV parameters, simulation count $S$, horizon $H$, block size $b$, quantile levels $Q$  
    \ENSURE Prediction intervals for monthly $y_{T+1:T+H}$, annual totals, and growth rates  
    \STATE Preprocess $y_{1:T}$.  
    \STATE Fit an algorithm to $y_{1:T}$.  
    \STATE Perform expanding-window CV to collect residual innovations $\{\hat{\varepsilon}_{k,h}\}$.  
    \STATE Center each horizon-wise residual column and extract all valid consecutive residual blocks of length $b$.  
    \FOR{$s=1$ {\bf to} $S$}  
        \STATE Sample residual blocks with replacement until length $H$, truncate to $H$, and add them to the forecast path to obtain $\widehat{y}_{s,T+1:T+H}$.  
        \STATE Aggregate annual totals: $\widehat{Y}_{s,year} = \sum_{m=1}^{12} \widehat{y}_{s,(year,m)}$ for forecast years.  
        \STATE Compute growth rates $\widehat{G}_{s,year} = (\widehat{Y}_{s,year} - \widehat{Y}_{s,year-1}) / \widehat{Y}_{s,year-1}$, with $\widehat{Y}_{s,year-1}$ initialized from last known annual sales in training.  
    \ENDFOR  
    \FOR{each target (monthly/annual/growth) and each quantile level $q \in Q$}  
        \STATE Compute empirical quantiles across $s=1,\ldots,S$ to form lower/upper bounds.  
    \ENDFOR  
    \STATE \textbf{Return} intervals and summary statistics (coverage and width).  
  \end{algorithmic}  
\end{algorithm}  
  
We compute empirical coverage rates and interval widths at multiple confidence levels (e.g., 90\%, 95\%, 99\%) across series and forecast scenarios (original, aggregated, growthrate). We assess statistical significance of differences versus a simulated-path baseline using Wilcoxon signed-rank tests.

\section{Experiments}  
We fit Auto-ARIMA models using the ARIMA implementation in the fable package \cite{fable2024}. Future trajectories are simulated using a modified version of the package’s \textit{generate()} function with block-bootstrap innovations drawn from cross-validated residual blocks. For each time series, we generate \(S = 10{,}000\) simulated paths.

We evaluate \SCP{} on M4 monthly sales data for aggregated annual totals (Aggregated Sales) and year-over-year growth rates (Y-o-Y Sales Growth), which are the primary targets of interest, and also report original monthly forecasts (Raw Sales) for reference. We compare against a baseline simulated-path approach (Baseline) \citep{hyndmanbook} that uses conditional simulation without cross-validated residual calibration. 

A direct split-conformal baseline built on aggregated nonconformity scores is less suitable in our setting because we start from monthly series and evaluate annual aggregates: once the data are aggregated to the yearly level, only a small number of calibration points remain for each series, making direct SCP at the aggregated level statistically inefficient and unstable.  

\begin{table*}[htbp]  
\centering  
\caption{Coverage on M4: \SCP{} vs Baseline}  
\label{tab:m4-coverage}  
\begin{tabular}{lcccccc}  
\toprule  
 & \multicolumn{3}{c}{\SCP{}} & \multicolumn{3}{c}{Baseline} \\  
Variant & 10\% & 5\% & 1\% & 10\% & 5\% & 1\% \\  
\midrule  
Raw Sales   & 88.8\% & 91.4\% & 93.9\% & 75.2\% & 80.8\% & 87.0\% \\  
Aggregated Sales & 83.1\% & 85.8\% & 88.9\% & 65.6\% & 70.9\% & 78.4\% \\  
Y-o-Y Sales Growth & 89.9\% & 92.0\% & 94.4\% & 75.2\% & 80.5\% & 87.4\% \\  
\bottomrule  
\end{tabular}  
\end{table*}  
  
\begin{table*}[htbp]  
\centering  
\caption{Interval widths on M4: \SCP{} vs Baseline}  
\label{tab:m4-widths}  
\begin{tabular}{lcccccc}  
\toprule  
 & \multicolumn{3}{c}{\SCP{}} & \multicolumn{3}{c}{Baseline} \\  
Variant & 10\% & 5\% & 1\% & 10\% & 5\% & 1\% \\  
\midrule  
Raw Sales   & $5.6 \times 10^{3}$ & $6.4 \times 10^{3}$ & $7.6 \times 10^{3}$ & $2.4 \times 10^{3}$ & $2.8 \times 10^{3}$ & $3.6 \times 10^{3}$ \\  
Aggregated Sales & $5.5 \times 10^{4}$ & $6.4 \times 10^{4}$ & $7.5 \times 10^{4}$ & $1.9 \times 10^{4}$ & $2.2 \times 10^{4}$ & $2.9 \times 10^{4}$ \\  
Y-o-Y Sales Growth & $1.6 \times 10^{0}$ & $2.3 \times 10^{0}$ & $6.6 \times 10^{0}$ & $4.5 \times 10^{-1}$ & $5.4 \times 10^{-1}$ & $7.3 \times 10^{-1}$ \\  
\bottomrule  
\end{tabular}  
\end{table*}  
    
\begin{table*}[htbp]  
\centering  
\caption{Coverage improvements of \SCP{} over Baseline on M4 data.}
\label{tab:m4-growth}  
\begin{tabular}{lccc}  
\toprule  
Dataset & Level & Coverage Delta & Coverage Cost \\  
\midrule  
Raw Sales   & 10\% & 13.6\% & 10.1 \\  
Raw Sales   & 5\%  & 10.7\% & 11.8 \\  
Raw Sales   & 1\%  & 6.9\% & 16.0 \\  
Aggregated Sales & 10\% & 17.5\% & 10.8 \\  
Aggregated Sales & 5\%  & 15.0\% & 12.3 \\  
Aggregated Sales & 1\%  & 10.4\% & 14.7 \\  
Y-o-Y Sales Growth & 10\% & 14.7\% & 17.2 \\ 
Y-o-Y Sales Growth & 5\%  & 11.5\% & 28.6 \\ 
Y-o-Y Sales Growth & 1\%  & 7.1\% & 113.1 \\  
\bottomrule  
\end{tabular}  
\end{table*}  
  
\begin{table*}[htbp]  
\centering  
\caption{Coverage on proprietary sales data: \SCP{} vs Baseline}  
\label{tab:sales-coverage}  
\begin{tabular}{lcccccc}  
\toprule  
 & \multicolumn{3}{c}{\SCP{}} & \multicolumn{3}{c}{Baseline} \\  
Variant & 10\% & 5\% & 1\% & 10\% & 5\% & 1\% \\  
\midrule  
Raw Sales   & 92.7\% & 94.5\% & 96.1\% & 82.8\% & 87.2\% & 91.7\% \\  
Aggregated Sales & 89.3\% & 91.1\% & 94.2\% & 75.3\% & 80.4\% & 86.8\% \\  
Y-o-Y Sales Growth & 89.3\% & 91.1\% & 94.2\% & 75.3\% & 80.4\% & 86.8\% \\  
\bottomrule  
\end{tabular}  
\end{table*}  
  
\begin{table*}[htbp]  
\centering  
\caption{Interval widths on proprietary sales data: \SCP{} vs Baseline}  
\label{tab:sales-widths}  
\begin{tabular}{lcccccc}  
\toprule  
 & \multicolumn{3}{c}{\SCP{}} & \multicolumn{3}{c}{Baseline} \\  
Variant & 10\% & 5\% & 1\% & 10\% & 5\% & 1\% \\  
\midrule  
Raw Sales   & $5.1 \times 10^{5}$ & $5.9 \times 10^{5}$ & $7.1 \times 10^{5}$ & $3.2 \times 10^{5}$ & $3.8 \times 10^{5}$ & $4.9 \times 10^{5}$ \\  
Aggregated Sales & $4.1 \times 10^{6}$ & $4.8 \times 10^{6}$ & $6.1 \times 10^{6}$ & $2.2 \times 10^{6}$ & $2.7 \times 10^{6}$ & $3.5 \times 10^{6}$ \\  
Y-o-Y Sales Growth & $1.4 \times 10^{0}$ & $1.6 \times 10^{0}$ & $2.1 \times 10^{0}$ & $5.9 \times 10^{-1}$ & $7.1 \times 10^{-1}$ & $9.4 \times 10^{-1}$ \\  
\bottomrule  
\end{tabular}  
\end{table*}  
    
\begin{table*}[htbp]  
\centering  
\caption{Coverage improvements of \SCP{} over Baseline on proprietary sales data.}
\label{tab:sales-growth}  
\begin{tabular}{lccc}  
\toprule  
Dataset & Level & Coverage Delta & Coverage Cost \\  
\midrule  
Raw Sales   & 10\% & 9.8\% & 6.3 \\  
Raw Sales   & 5\%  & 7.3\% & 7.5 \\  
Raw Sales   & 1\%  & 4.4\% & 10.2 \\  
Aggregated Sales & 10\% & 14.0\% & 5.9 \\  
Aggregated Sales & 5\%  & 10.7\%  & 7.2 \\  
Aggregated Sales & 1\%  & 7.4\%  & 9.7 \\  
Y-o-Y Sales Growth & 10\% & 14.0\% & 9.4 \\  
Y-o-Y Sales Growth & 5\%  & 10.7\%  & 12.2 \\  
Y-o-Y Sales Growth & 1\%  & 7.4\%  & 17.1 \\  
\bottomrule  
\end{tabular}  
\end{table*}  

Across all targets and levels, \SCP{} delivers higher coverage than Baseline with a corresponding increase in interval width, but both methods miss the target coverage level. The coverage and width values are presented in Tables~\ref{tab:m4-coverage} and \ref{tab:m4-widths}, respectively. In absolute terms, coverage gains range from roughly 6.9 to 17.5 percentage points (Column \textit{Coverage Delta} in Table~\ref{tab:m4-growth}). %\paragraph{Trade-off analysis.}  
Relative to Baseline, \SCP{} improves coverage while widening intervals. The coverage cost summarizes the relative width increase per coverage gain: costs are lowest at the 10\% level, increase at 5\%, and are highest at 1\% (Column \textit{Coverage Cost} in Table~\ref{tab:m4-growth}). Raw Sales exhibits the lowest cost across levels, whereas Y-o-Y Sales Growth is the most costly—especially at 1\%. These wider intervals are expected under compounded uncertainty from temporal aggregation.  
  
\paragraph{Real-world monthly sales.}  
We additionally evaluate on 2{,}000 actual monthly sales series. As with M4, \SCP{} attains higher coverage with wider intervals at higher confidence. For Raw Sales, \SCP{} achieves 92.7\%, 94.5\%, and 96.1\% coverage at the 10\%, 5\%, and 1\% levels, respectively. This performance is better than on M4, but it still undercovers except at the 10\% miscoverage level. Aggregated Sales and Y-o-Y Sales Growth exhibit slightly lower coverages but the same qualitative pattern. In Table~\ref{tab:sales-coverage}, Aggregated Sales and Y-o-Y Sales Growth have identical coverage values because the proprietary test horizon is only 12 months, so annual aggregation produces a single forecast-year total. The corresponding growth rate is then obtained by dividing each simulated annual sum by the same observed annual sum from the previous year. This is a monotone transformation, so if an interval for aggregated sales covers the annual sum, then the corresponding growth-rate interval also covers the associated growth rate; conversely, if the aggregated-sales interval misses the annual sum, then the growth-rate interval also misses the associated growth rate. 

Relative to Baseline, \SCP{} improves coverage by 4.4--14.0 percentage points (Column \textit{Coverage Delta} in Table~\ref{tab:sales-growth}). The coverage cost is smallest for Aggregated Sales at 10\% and increases as nominal levels tighten; the largest cost occurs at the 1\% level for Y-o-Y Sales Growth, which is lower in magnitude than on M4 but shows the same qualitative trend.  
    
Overall, \SCP{} achieves materially higher coverage than Baseline across both M4 and proprietary real-world sales data, at the expense of wider intervals. On M4, the width increase per unit coverage gain is smallest for Aggregated at 10\% and largest for Growthrate at 1\%. On the proprietary dataset, the corresponding extremes are likewise smallest for Aggregated at 10\% and largest at 1\%, consistent with aggregation effects observed in both settings.  

In the absence of exchangeability, conformal prediction methods do not enjoy finite-sample coverage guarantees, and nominal levels should therefore be interpreted as target rather than guaranteed coverage. Consistent with this theoretical limitation, our empirical results show that while \SCP{} substantially improves coverage relative to simulated-path baselines, achieved coverage remains below nominal levels as shown in Tables~\ref{tab:m4-coverage} and \ref{tab:sales-coverage}. This behavior reflects the intrinsic difficulty of uncertainty quantification in non-stationary and dependent time-series settings, rather than a failure of the proposed method.

In practice, users requiring closer alignment between nominal and achieved coverage may apply post-hoc calibration strategies, such as increasing the nominal coverage level, or applying online adaptation schemes introduced in \citep{Gibbs2021}. Investigating principled calibration schemes that account for temporal dependence is an important direction for future work.

\paragraph{Significance testing.}  
Wilcoxon signed-rank tests comparing \SCP{} vs Baseline yield $p$-values $\ll 0.001$ for coverage and width across Aggregated and Growthrate targets at 10\%, 5\%, and 1\% levels. Differences are statistically significant.

\section{Conclusion}  
We propose \SCP{}, a simulation-enhanced multi-step split conformal method that combines expanding-window cross-validation, block-bootstrap future-path simulation, and calibration on aggregated trajectories. \SCP{} improves empirical coverage for monthly, aggregated annual, and growth-rate targets, at the cost of wider intervals, reflecting a fundamental trade-off in conformal prediction under dependence and aggregation. Residuals across forecast origins and horizons are not strictly exchangeable; expanding-window cross-validation partially accounts for temporal dependence by pooling residuals over time, while the block bootstrap preserves local cross-horizon dependence in simulated paths. Sequential CP methods (e.g., SPCI, PID) target pointwise horizons and long-run coverage, making extension to aggregated quantities nontrivial. Copula-based calibration could model cross-horizon dependence for sums and growth rates, but requires multiple time series, whereas our setting focuses on a single series \citep{SunYu2022}.  

Overall, \SCP{} provides a practical and scalable approach to uncertainty quantification for aggregated forecasting tasks.

\bibliography{example_paper}
\bibliographystyle{icml2026}

\end{document}